\pdfoutput=1

\documentclass[11pt]{article}

\usepackage[]{ACL2023}

\usepackage{times}
\usepackage{latexsym}

\usepackage{graphicx}
\usepackage{enumitem}
\usepackage{caption}
\usepackage{multirow}
\usepackage{array}
\usepackage{subcaption}
\usepackage{dirtytalk}
\usepackage{xspace}
\usepackage{booktabs}
\usepackage{amsmath}
\usepackage{amssymb}
\usepackage{balance}
\usepackage{fixltx2e}
\usepackage{diagbox}

\newcommand{\ie}{\emph{i.e.,}\xspace}
\newcommand{\eg}{\emph{e.g.,}\xspace}

\newcommand{\mname}{Chunk-OIE\xspace}

\usepackage[T1]{fontenc}

\usepackage[utf8]{inputenc}

\usepackage{microtype}

\usepackage{inconsolata}
\usepackage[normalem]{ulem}

\title{Open Information Extraction via Chunks}

\author{Kuicai Dong$^{1,2}$, Aixin Sun$^1$, Jung-Jae Kim$^2$, Xiaoli Li$^{1,2,3}$ \\
\fontsize{11pt}{12pt}\selectfont
$^1$ School of Computer Science and Engineering, Nanyang Technological University, Singapore\\
\fontsize{11pt}{12pt}\selectfont
\texttt{kuicai001@e.ntu.edu.sg, axsun@ntu.edu.sg}\\
\fontsize{11pt}{12pt}\selectfont
$^2$ Institute for Infocomm Research, A*STAR, Singapore\\
\fontsize{11pt}{12pt}\selectfont
$^3$ A*STAR Centre for Frontier AI Research, Singapore\\
\fontsize{11pt}{12pt}\selectfont
\texttt{\{jjkim, xlli\}@i2r.a-star.edu.sg}}

\begin{document}
\maketitle
\begin{abstract}

Open Information Extraction (OIE) aims to extract relational tuples from open-domain sentences. Existing OIE systems split a sentence into tokens and recognize token spans as tuple relations and arguments. We instead propose 
\textit{\textbf{S}entence \textbf{a}s \textbf{C}hunk sequence} (\textbf{SaC}) and recognize chunk spans as tuple relations and arguments.
We argue that SaC has better quantitative and qualitative properties for OIE than sentence as token sequence, and evaluate four choices of chunks (\ie CoNLL chunks, simple phrases, NP chunks, and spans from SpanOIE) against gold OIE tuples. 
Accordingly, we propose a simple BERT-based model for sentence chunking, and propose \mname for tuple extraction on top of SaC. 
\mname achieves state-of-the-art results on multiple OIE datasets, showing that SaC benefits OIE task. Our model will be publicly available in Github\footnote{https://github.com/daviddongkc/Chunk\_OIE}.

\end{abstract}

\section{Introduction}\label{sec:intro}

Open Information Extraction (OIE or OpenIE) is to extract structured tuples from unstructured open-domain text~\cite{yates2007textrunner}. The extracted tuples are in the form of (\textit{Subject}, \textit{Relation}, \textit{Object}) if binary relations, and (\textit{ARG}\textsubscript{0}, \textit{Relation}, \textit{ARG}\textsubscript{1}, \dots, \textit{ARG}\textsubscript{n}) $n$-ary relations. The structured relational tuples are beneficial to many downstream tasks such as question answering~\cite{khot2017answering} and knowledge base population~\cite{martinez2018openie, gashteovski2020aligning}. 

Observe from benchmark OIE datasets, most relations and their arguments are \textit{token spans}. 
As a domain-independent information extraction task, OIE does not specify any pre-defined extraction schema. Hence, the granularity or length of such text spans is hard to define. 
Consequently, many of existing OIE systems extract tuples at the \textit{token level}, tagging every token in a sentence by using BIO\footnote{Begin, Inside, and Outside of a subject/relation/object.} or a similar tagging scheme, called tagging-based methods. 
By contrast, generative methods can generate any tokens as OIE tuples.

Recently, \citet{sun-etal-2020-predicate} and \citet{wang-etal-2022-oie} propose to use Open Information Annotation (OIA) as an intermediate layer between the input sentence and  OIE tuples. The OIA of a sentence is a graph where nodes are simple phrases, and edges connect the predicate nodes and their argument nodes. 
The OIA graphs can be converted into OIE tuples by using dataset-specific rules. However, it is challenging to correctly generate the entire OIA graph of a given sentence.

\begin{figure}
    \centering
    \begin{subfigure}[b]{\columnwidth}
         \includegraphics[width=\columnwidth]{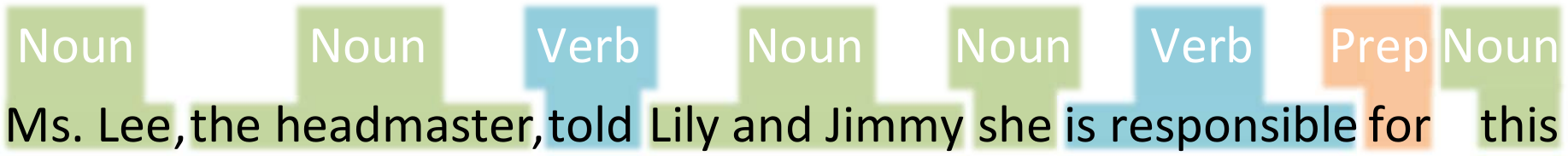}
         \caption{OIA simple phrases}
         \label{subfig:oia_chunk}
     \end{subfigure}
  \begin{subfigure}[b]{\columnwidth}
         \includegraphics[width=\columnwidth]{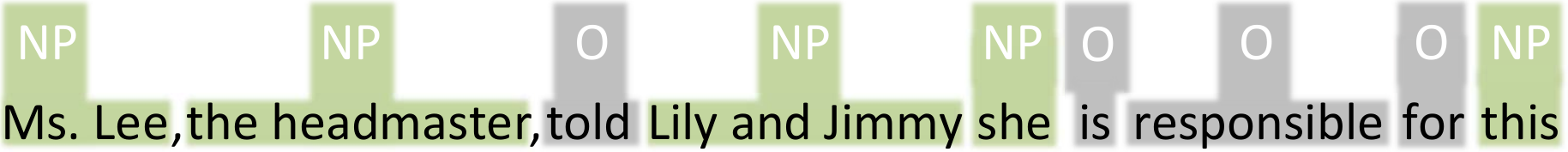}
         \caption{Noun phrases (NPs)}
         \label{subfig:np_chunk}
     \end{subfigure}
  \begin{subfigure}[b]{\columnwidth}
         \includegraphics[width=\columnwidth]{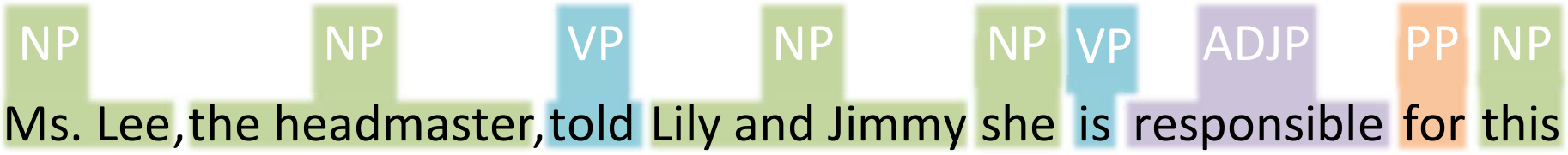}
         \caption{CoNLL-chunked phrases}
         \label{subfig:conll_chunk}
     \end{subfigure}
    \caption{A sentence in different chunk sequences. }
    \label{fig:sent_chunk}
    
\end{figure}

Inspired by OIA, we propose the notion of \textit{\textbf{S}entence \textbf{a}s \textbf{C}hunk sequence} (\textbf{SaC}), as an alternative intermediate layer representation. 
Chunking is a type of shallow parsing, dividing a sentence into syntactically related non-overlapping phrases, called chunks~\cite{tjong-kim-sang-buchholz-2000-introduction}. For instance, the simple phrases in OIA can be considered as chunks (Figure~\ref{subfig:oia_chunk}). 
To justify the adaptability of SaC for OIE, we also employ other choices of chunks, including Noun Phrase chunks (Figure~\ref{subfig:np_chunk}) and CoNLL chunks (Figure~\ref{subfig:conll_chunk}).
Figure~\ref{fig:sent_chunk} shows an example sentence with different kinds of chunking scheme. 
We then propose \mname, a tagging-based neural OIE model with two sub-models for two subtasks: (i) to represent sentence in SaC, and (ii) to extract tuples based on SaC. We show that SaC significantly outperforms the conventional notion of Sentence as Token sequence adopted by most of neural OIE methods, if OIE tuple relations and arguments align well with the chunks, which is often the case.

Our contributions are the followings. 
We propose a novel notion of Sentence as Chunk sequence (SaC) for OIE. We compare SaC and OIA from the perspectives of feasibility, flexibility, and adaptability for OIE tasks ($\S$~\ref{ssec:sacVsOia}). Through data analysis ($\S$~\ref{ssec:boundary}) against gold tuples, we show that chunks have a suitable granularity of token spans for OIE. We design a strategy to capture dependencies at chunk level, largely simplifying the token-level dependencies.
We propose \mname, with two  sub-models to (i) represent sentence as SaC ($\S$~\ref{sec:sent_chunk}), and (ii) extract tuples based on SaC ($\S$~\ref{sec:oie_model}). Experiment results show the effectiveness of \mname against strong baselines. We believe \mname represents a novel way of approaching OIE.

\section{Related Work}

\paragraph{OIE Systems.}

OIE was first proposed by \citet{yates2007textrunner}, and TextRunner is the first system that generates relational tuples in open domain. 
Before deep learning era, many statistical and rule-based systems have been proposed, including Reverb~\cite{fader2011identifying}, OLLIE~\cite{schmitz2012open}, Clausie~\cite{del2013clausie}, Stanford OIE~\cite{angeli2015leveraging}, OpenIE4~\cite{mausam2016open}, and MINIE~\cite{gashteovski2017minie}. 
These models extract relational tuples typically based on syntactic structures such as part-of-speech (POS) tags and dependency trees.

Recently, two kinds of neural systems have been explored, generative and tagging-based systems~\cite{ijcai2022p793}. 
Generative OIE systems~\cite{cui2018neural, kolluru2020imojie, dong2021docoie} model tuple extraction as a sequence-to-sequence generation task with copying mechanism. 
Tagging-based OIE systems~\cite{stanovsky2018supervised, kolluru2020openie6, kotnis-etal-2022-milie} tag each token as a sequence labeling task.
SpanOIE~\cite{zhan2020span} uses a different approach. It enumerates all possible spans (up to a predefined length) from a sentence. After rule-based filtering, the remaining candidate spans are classified to relation, argument, or not part of a tuple. However, enumerating and filtering all possible spans for scoring is computational expensive.

Early neural models typically do not utilize syntactic structure of sentence, which was required by traditional models. Recently works show that encoding explicit syntactic information  benefits neural OIE as well. RnnOIE~\cite{stanovsky2018supervised} and SenseOIE~\cite{roy2019supervising} encode POS / dependency as additional embedding features.
MGD-GNN~\cite{MGD-GNN_2021} connects words, if they are in dependency relations, in an undirected graph and applies GAT as its graph encoder. RobustOIE~\cite{qi-etal-2022-syntactically} uses paraphrases (with various constituency form) for more syntactically robust OIE training.
SMiLe-OIE~\cite{dong2022smile_oie} incorporates heterogeneous syntactic information (constituency and dependency graphs) through GCN encoders and multi-view learning.
Inspired by them, we design a simple strategy to model dependency relation at the chunk level. Note that chunks in SaC partially reflect constituency structure as words in a chunk are syntactically related, by definition.

\paragraph{Sentence Chunking.} Our proposed notion of SaC is based on the concept of chunking. Chunking is to group tokens in a sentence into syntactically related non-overlapping groups of words, \ie chunks.
Sentence chunking is a well studied pre-processing step for sentence parsing. The earliest task of chunking was to recognize non-overlapping noun phrases \cite{ramshaw-marcus-1995-text} as exemplified in Figure~\ref{subfig:np_chunk}. Then CoNLL-2000 shared task~\cite{tjong-kim-sang-buchholz-2000-introduction} proposed to identify other types of chunks such as verb and prepositional phrases, see Figure~\ref{subfig:conll_chunk}.

\paragraph{OIX and OIA.}
\citet{sun-etal-2020-predicate} proposed Open Information eXpression (OIX) as a new pipeline to build OIE systems. The key idea of OIX is to represent a sentence in an intermediate layer without information loss, so that reusable OIE strategies can be developed on top of the latter. As an implementation, they propose Open Information Annotation (OIA) format. 
OIA of a sentence is a single-rooted directed-acyclic  graph (DAG). Its basic information unit, \ie graph node, is a simple phrase.
A simple phrase is either a fixed expression or a phrase. \citet{sun-etal-2020-predicate} define simple phrases with types like constant (\eg nominal phrase), predicate (\eg verbal phrase), and functional (\eg wh-phrase). 
Edges in an OIA graph connect the predicate nodes and function nodes to their arguments.
\citet{wang-etal-2022-oie} extend OIA by defining more simple phrase types and release an updated version of the OIA dataset. The authors also propose OIA@OIE, which first trains an OIA generator to produce OIA graphs, and then uses different rule-based OIE adaptors to extract tuples based on OIA graphs.
Note that OIE@OIA requires dedicated rule strategies for each OIE dataset, taking considerable manual efforts.

\section{Sentence as Chunk sequence (SaC)} \label{sec:chunk}
As its name suggests, SaC is to represent a sentence in \textit{syntactically related and non-overlapping} chunks. 
SaC can be realized with any chunking scheme. For instance, all the three chunk sequences in Figure~\ref{fig:sent_chunk} can be used.
In this section, we justify the effectiveness of SaC for OIE with the following analyses: 
(i) comparison between SaC and OIA as intermediate representations ($\S$~\ref{ssec:sacVsOia}), 
(ii) syntactical modelling of input sentence ($\S$~\ref{ssec:syntax}),
and (iii) alignment between boundaries of chunks and OIE relations/arguments ($\S$~\ref{ssec:boundary}).

\subsection{SaC versus OIA}
\label{ssec:sacVsOia}
We compare SaC and OIA as intermediate representations for OIE tasks from three qualitative perspectives: feasibility, flexibility, and adaptability.

\paragraph{Feasibility.} Sentence chunking is a classic NLP task and has been well studied. With Pretrained Language Models (PLM), sentence chunking can be achieved with very high accuracy, \eg $F_1$ score of 97.3 on CoNLL chunking task \cite{wang-etal-2021-automated}. In comparison, learning OIA graph consists of two subtasks. The first is to identify and classify simple phrases as nodes, which is a type of sentence chunking. The second is to establish edges between these nodes and to determine edge types to complete the OIA graph. The latest solution for OIA~\cite{wang-etal-2022-oie} achieves $F_1$ scores of 88.5\%, 69.8\%, 52.5\% at node, edge, and graph levels, respectively. Due to its complex design, it remains challenging to generate high-quality OIA graph from sentence.   

\paragraph{Flexibility.}
PLMs such as BERT~\cite{devlin2018bert} have shown promising results on many NLP tasks, through implicit yet effective semantic encoding of sentence.
SaC, as sequential chunks, can directly map to the input format of PLMs, fully leveraging the power of PLMs. The graph structure of OIA, on the other hand, is a type of explicit semantic encoding. Such kind of encoding enables rule-based adaptors to convert OIA graph into OIE tuples, as envisioned in~\cite{sun-etal-2020-predicate}. However, the OIA graph generation does not directly benefit from the sequential encodings by PLMs. Moreover, SaC can be easily adopted to other NLP tasks that involve phrase recognition such as corefenrence resolution and semantic role labelling. In contrast, OIA is purposely designed for OIE and extending OIA to other tasks would require additional effort. 

\paragraph{Adaptability.}
One key notion of OIX is to increase the adaptability to different OIE datasets with less cost.
However, there is no universal strategy to convert OIA into OIE tuples.
Hence,  OIE@OIA~\cite{wang-etal-2022-oie} hand-crafted conversion rules for each OIE dataset. Although the authors claim that some rule strategies are reusable, domain expertise and considerable manual efforts are required. 
With SaC, neural models can be learned in end-to-end manner for different OIE formats. Nevertheless, the learning of neural models requires data annotations, either gold or silver.

\begin{figure}
\centering
\begin{subfigure}[b]{1\linewidth}
   \centering
   \includegraphics[width=1\linewidth]{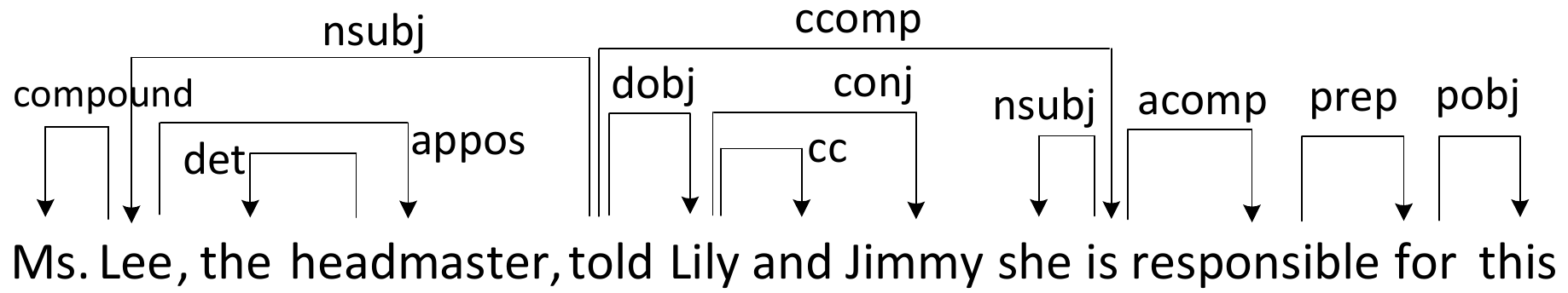}
   \caption{Dependency tree (results from spaCy).}
   \label{fig:dep_word}
\end{subfigure}
\begin{subfigure}[b]{1\linewidth}
   \centering
   \includegraphics[width=1\linewidth]{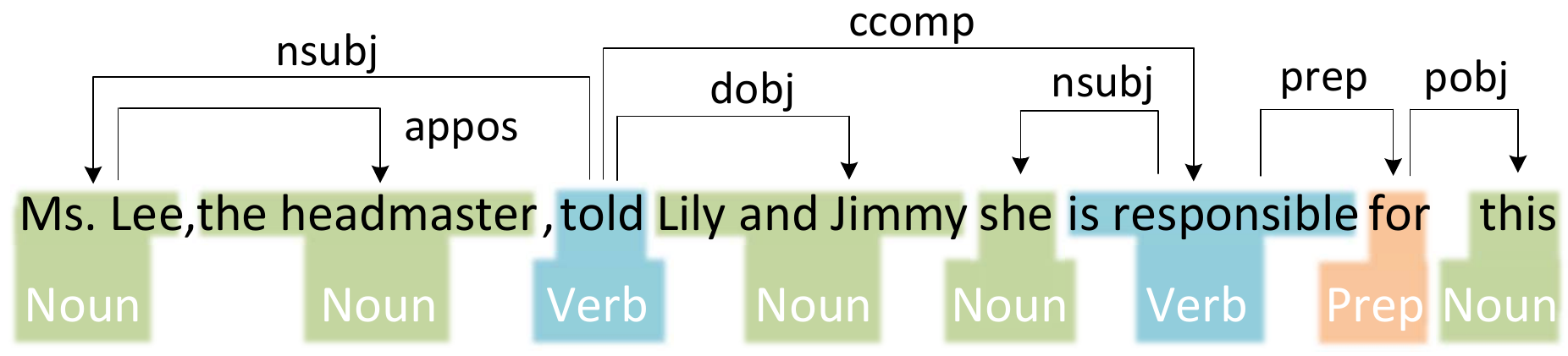}
   \caption{Dependency tree at chunk level with OIA-SP.}
   \label{fig:dep_phrase}
\end{subfigure}
\caption{Dependency trees at token-level and chunk-level (in OIA simple phrases), respectively.}
\label{fig:dep_example}
\end{figure}

\begin{table}[t] 
\small
  \begin{tabular}{l|l|l}
    \toprule
    \multicolumn{2}{c|}{Scenario}  &  Example\\
    \midrule
    \multirow{2}{*}{Match}  & Exact &  \begin{minipage}{.225\textwidth}
      \includegraphics[width=\linewidth]{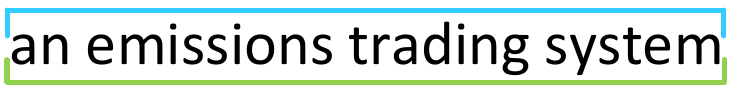}
    \end{minipage}\\ 
      & Concatenation &   \begin{minipage}{.195\textwidth}
      \includegraphics[width=\linewidth]{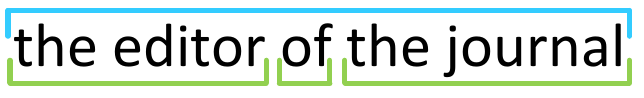}
    \end{minipage}\\   
    \midrule
    \multirow{2}{*}{Mismatch} & Overlap &  \begin{minipage}{.21\textwidth}
      \includegraphics[width=\linewidth]{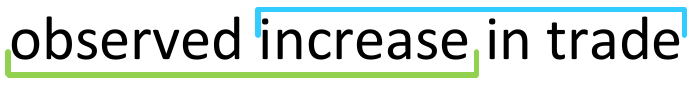}
    \end{minipage}\\
    & NoOverlap & \begin{minipage}{.20\textwidth}
      \includegraphics[width=\linewidth]{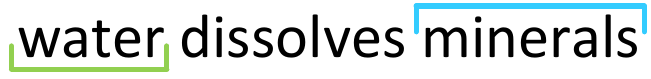}
    \end{minipage}\\
  \bottomrule
  \end{tabular}
\caption{Four scenarios for matching a gold tuple span (boundary marked in blue) to a generated chunk (boundary marked in green).}
\vspace{-1.5em}
\label{tab:span_case}
\end{table}

\begin{table*}[t] 
\small
    \centering
  \begin{tabular}{ll|rr|rr|rr|rr|rr}
    \toprule
    \multicolumn{2}{c|}{\multirow{2}{*}{\backslashbox{Match Case}{Chunk}}}   &  \multicolumn{2}{c|}{CoNLL} & \multicolumn{2}{c|}{OIA-SP} & \multicolumn{2}{c|}{NP\textsubscript{short}} & \multicolumn{2}{c|}{NP\textsubscript{long}} & \multicolumn{2}{c}{SpanOIE}\\ 
     & & Percent & $L_p$ &  Percent & $L_p$ & Percent & $L_p$ & Percent & $L_p$ & Percent & $L_p$ \\
    \midrule
    \parbox[t]{2mm}{\multirow{4}{*}{\rotatebox[origin=c]{90}{Precision}}} 
    &Match  & \textbf{51.0\%} & 1.8 & \underline{49.7\%} & 2.0 & 49.0\% & 1.7 & 40.5\% & 2.3 &  3.3\% & 3.5\\
     &\quad\quad-Exact & 8.4\% & 2.3 & \underline{10.2\%} & 2.5 & 7.2\% & 2.1 & \textbf{11.0\%} & 3.4 &  3.3\% &  3.5 \\
     &\quad\quad-Concatenation & \textbf{42.6\%} & 1.7 & 39.5\% & 1.9  & \underline{41.8\%} & 1.6 & 29.5\% & 1.9 &  -  &  - \\
     & Mismatch-NoOverlap & 49.0\% & 1.4 & 50.3\% & 1.6 & 51.0\% & 1.4 & \underline{59.5\%} & 2.3 &  \textbf{96.7\%} & 4.4  \\ 
    \midrule
     & Matching case & Percent & $L_s$ &  Percent & $L_s$ & Percent & $L_s$ & Percent & $L_s$ & Percent & $L_s$ \\
     \midrule
     \parbox[t]{2mm}{\multirow{4}{*}{\rotatebox[origin=c]{90}{Recall}}}
    &Match  & \textbf{90.5\%} & 4.4 & \underline{89.9\%} & 4.5 & 79.7\% & 4.3 & 58.7\% & 4.7 &  86.0\% & 3.3\\
     &\quad\quad-Exact  & \underline{45.7}\% & 2.3 & \textbf{48.9\%} & 2.5 & 37.1\% & 2.1 & 36.7\% & 3.4 &  86.0\% & 3.3  \\
     &\quad\quad-Concatenation  & \textbf{44.8\%} & 6.4 & 41.0\% & 6.8 & \underline{42.6\%} & 6.2 & 22.0\% & 7.1 &  - & -  \\
    &Mismatch-Overlap  & 9.5\% & 4.3 & 10.1\% & 3.6 & \underline{20.3\%} & 4.4 & \textbf{41.3\%} & 4.1 &  14.0\% & 12.7  \\ 
  \bottomrule
  \end{tabular}
  \vspace{-0.5em}
\caption{Precision and Recall Analysis. $L_{s}$ and $L_{p}$ are length of gold spans and generated chunks, respectively. For each type of match/mismatch case, the highest score is in boldface and second highest score is underlined.
\vspace{-1.0em}
}
\label{tab:span_analysis}
\end{table*}

\begin{table}[t]
\centering
  \small
  \begin{tabular}{l|rrr}
    \toprule
    Candidate chunks  & Precision & Recall & $F_1$ \\
    \midrule
    CoNLL & \textbf{51.0} & \textbf{90.5} & \textbf{65.2} \\
    OIA-SP & \underline{49.7} & \underline{89.9} & \underline{64.0} \\
    NP\textsubscript{short} & 49.0 & 79.7 & 60.7\\
    NP\textsubscript{long}   & 40.5 & 58.7 & 47.9 \\
    SpanOIE &  3.3  &  86.0 &  6.4 \\

  \bottomrule
  \end{tabular}
  \vspace{-0.5em}
\caption{Precision, Recall, and $F_1$ of generated chunks; best scores are in boldface, second best underlined.}
\vspace{-1.0em}
\label{tab:span_f1}
\end{table}

\subsection{Syntactic Modelling Analysis} \label{ssec:syntax}
By grouping words into typed chunks, \textit{SaC greatly simplifies the modeling of sentence syntactical structure} for OIE. Recent studies show that both constituency and dependency structures benefit neural models for information extraction tasks including OIE~\cite{fei-etal-2021-better,dong2022smile_oie}. 
However, directly incorporating both constituency and dependency relationships in neural models is complicated. On the other hand, the typed chunks in SaC well reflect the constituency structure at phrasal level. Further, the SaC representation largely simplifies the dependency structure, which in turn facilitates OIE (see Appendix~\ref{ssec:dep_chunk} about the simplification and facilitation). Figure~\ref{fig:dep_example} shows the dependency tree on SaC (with OIA simple phrase as chunks) verses dependency tree at word level.

\subsection{Chunk Boundary Analysis} \label{ssec:boundary}
We now perform boundary alignment analysis of SaC against gold spans in a benchmark OIE dataset named LSOIE.
Gold Spans are the token spans of tuple arguments / relations in ground truth annotations.
We analyze CoNLL chunks, OIA simple phrases, and NP chunks as chunk choices for SaC. We also include the enumerated spans of SpanOIE in our analysis. Detailed description and statistics of these chunks are in \ref{ssec:chunk_stats}.
The  boundary alignment analysis is conducted from two perspectives. (i)  Precision: How often do the boundaries of SaC chunks match those of gold spans?  (2)  Recall: How often do the boundaries of gold spans match those of SaC chunks? 
There are four scenarios of boundary alignment, as exemplified in Table~\ref{tab:span_case}.
\textbf{Match-Exact}: A gold span can be exactly matched to a chunk. 
\textbf{Match-Concatenation}: A gold span can be mapped to multiple chunks in a consecutive sequence.\footnote{This is not applicable to SpanOIE, since it emulates all possible spans; if there is a match, it should be an exact match.}
\textbf{Mismatch-Overlap}: A chunk overlaps with a gold span, and at least one token of the chunk is not in the gold span.
\textbf{Mismatch-NoOverlap}: A chunk does not overlap with any gold span.

We show the precision and recall analysis of four boundary alignment scenarios in Table~\ref{tab:span_analysis} and summarize the overall scores in Table~\ref{tab:span_f1}. 
Observe that CoNLL chunks and OIA simple phrases show higher precision and recall of the Match boundary alignment than the other chunks.
We also note that SpanOIE has only 3.3\% of precision, indicating that enumerating all possible spans should bear heavy burden to detect correct spans.

\section{\mname Model}

\begin{figure*}
    \centering
    \begin{subfigure}{0.5\textwidth}
        \centering
        \includegraphics[width=0.95\columnwidth]{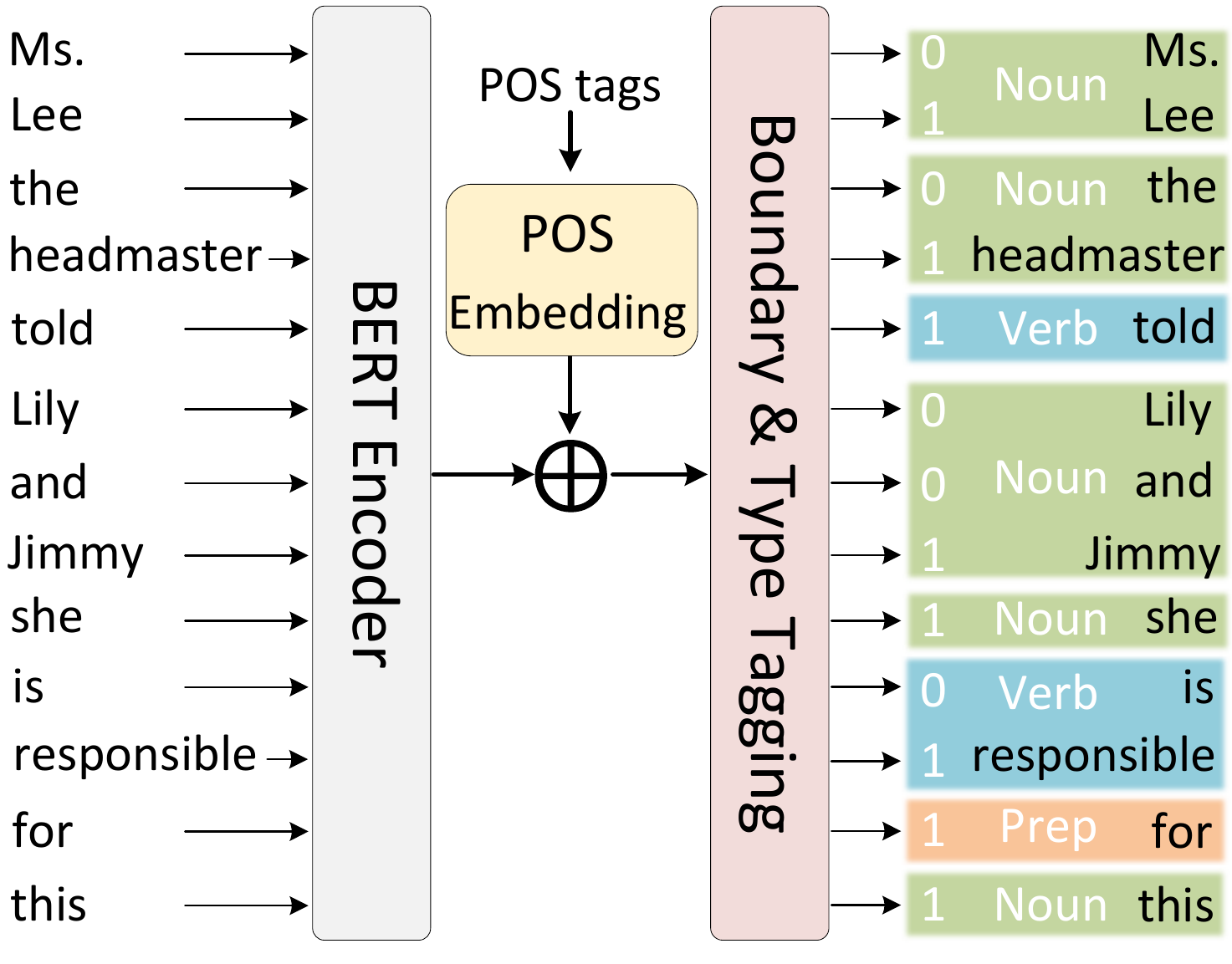}
        \caption{Representing Sentence as Chunks (SaC)}
        \label{fig:oie_modela}
    \end{subfigure}%
    ~ 
    \begin{subfigure}{0.5\textwidth}
        \centering
        \includegraphics[width=0.95\columnwidth]{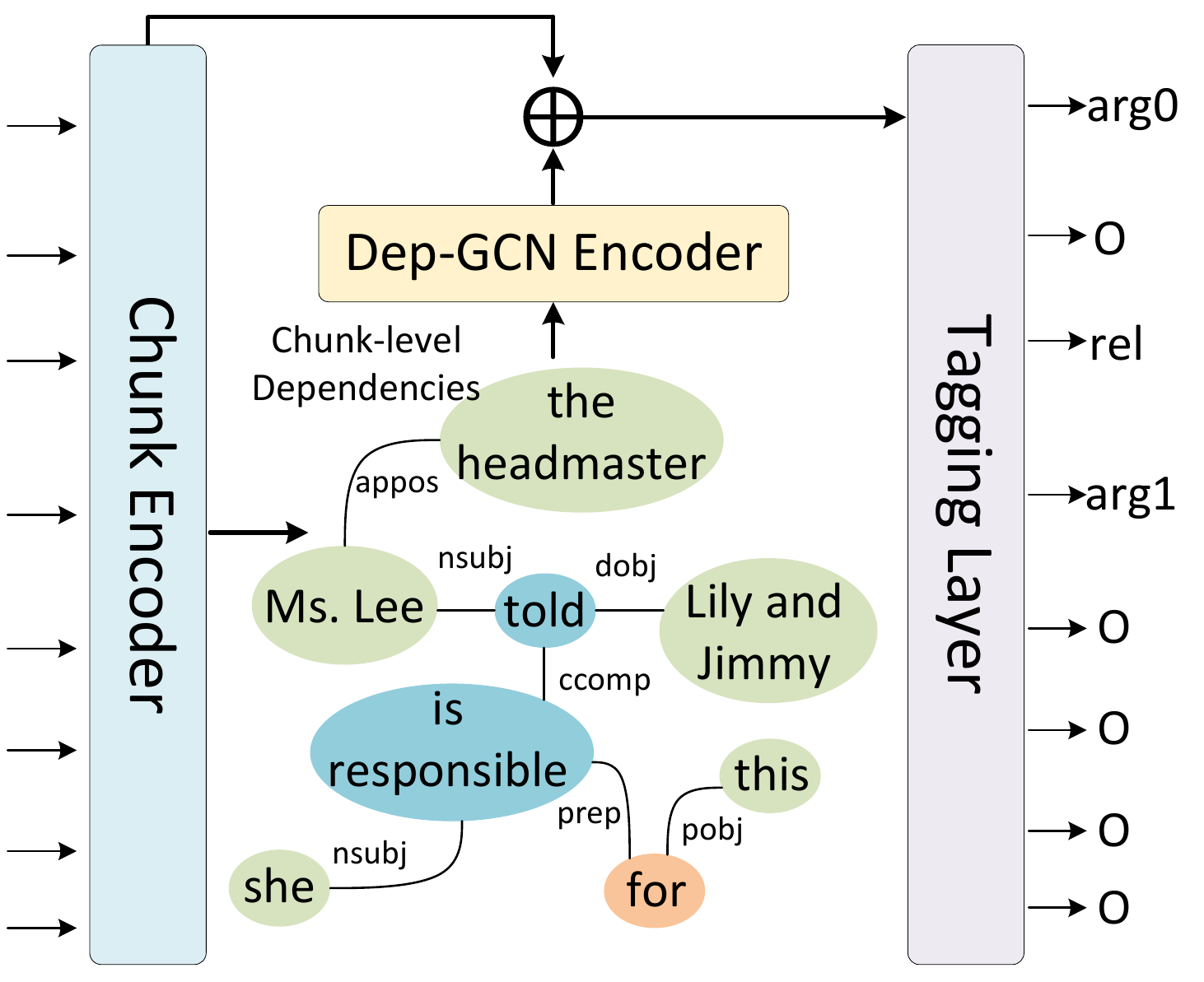}
        \caption{SaC‐based OIE}
        \label{fig:oie_modelb}
    \end{subfigure}
    \caption{The overview of \mname. Punctuation marks in the sentence are neglected for conciseness. \mname consists of two stages: (i)  representing Sentence as Chunks (SaC) in Figure~\ref{fig:oie_modela}, (ii) SaC-based OIE tuple extraction in Figure~\ref{fig:oie_modelb}. The SaC part is pre-trained with chunking dataset and frozen during the training of OIE tuple extraction.}
    
    \vspace{-1em}
    \label{fig:oie_model}
\end{figure*}

\subsection{Representing Sentence as Chunks (SaC) }
\label{sec:sent_chunk}
We formulate SaC (see Figure~\ref{fig:oie_modela}) as two sequence tagging sub-tasks: (i) binary classification for chunk boundary, and (ii) multi-class classification for chunk type. We address both sub-tasks via  multi-task learning. Tokens at boundaries are tagged as 1 and non-boundaries as 0.

Specifically, we first use BERT to get the contextual representations of input tokens $[t_1, \dots, t_n]$.
Subsequently, we generate POS representations with a trainable embedding matrix $\boldsymbol{W_{\mathrm{POS}}}$.
We obtain the hidden representations of tokens by concatenating the corresponding BERT and POS representations: $h_i^{'} = \boldsymbol{W_{\mathrm{BERT}}}(t_i)  +  \boldsymbol{W_{\mathrm{POS}}}(t_i) \in \mathbb{R} ^ {d_h}$.
$h_i^{'}$ is then passed into tagging layers for chunk boundary and type classification concurrently.

\paragraph{SaC Learning.}
Considering that the two sub-tasks (boundary and type classification) benefit each other, we combine their cross-entropy losses to jointly optimize our chunking model: 
\begin{equation}
    L_{bound} = - \sum^{n}_{i=1} y_i^b \mathrm{log}(p^b_i) + (1-y^b_i) \mathrm{log}(1-p^b_i)
\end{equation}

\begin{align}
    L_{type} &= - \sum^{n}_{i=1} \sum^c_{j=1} y_{i,j}^t \mathrm{log}(p_{i,j}^t)\\
 L_{chunk} &= L_{bound} + \alpha L_{type} 
 \label{eq:loss_sac}
\end{align}
Here, $y^b$ and $p^b$ are gold label and softmax probability for chunk boundary. $y^t$ and $p^t$ are gold label and softmax probability for chunk type.
$c$ refers to the number of chunk types. $\alpha$ is a hyperparameter balancing the two losses.

\subsection{SaC based OIE}
\label{sec:oie_model}
SaC-based OIE model corresponds to Figure~\ref{fig:oie_modelb}.
Given the chunk boundaries and types classified by the chunking model in Section \ref{sec:sent_chunk}, we convert the BERT token representations into chunk representations, and encode the chunk type embedding.
Subsequently, we model the sequential chunks into chunk-level dependency graph. Finally, we use Graph Convolution Network (GCN) to get the chunk-level dependency graph representations. The tagging layer performs tagging at the chunk-level to extract OIE tuples, based on the concatenated BERT-based and GCN-based chunk representations.

\paragraph{OIE Task Formulation.}
We formulate OIE as a sequence tagging task on top of chunks. That is, given the chunks of the input sentence, we perform the tagging on the chunk sequence $[c_1, \dots, c_m]$  rather than on the token sequence $[t_1, \dots, t_n]$. A variable number of tuples are extracted from a sentence. Each tuple can be represented as $[x_1, \dots, x_L]$, where each $x_i$  (\ie relation or argument) is a contiguous span of chunks, either an exact match or chunk concatenation. One of $x_j$ is a tuple relation (REL) and the others tuple are tuple arguments (ARG$_l$). For instance, the tuple in Figure~\ref{fig:oie_model} can be represented as $($arg$_0$=`Ms. Lee', rel=`told', arg$_1$=`Lily and Jimmy'$)$.

\paragraph{Chunk Encoder.} 
We employ BERT as encoder to get the token representations in a sentence. As each verb in a sentence is a potential relation indicator, verb embedding is useful to highlight this candidate relation indicator~\cite{dong2022smile_oie}. We follow \citet{dong2022smile_oie} to encode tokens with additional verb embeddings, \ie $w_i=\boldsymbol{W_{word}}(t_i)+\boldsymbol{W_{verb}}(t_i)$, where $\boldsymbol{W_{word}}$ is trainable and initialized by BERT word embedding, and $\boldsymbol{W_{verb}}$ is a trainable verb embedding matrix. Then, we input $w_i$ to the BERT encoder and utilize BERT's last hidden states as token-level representations $h_i^{token}$.

For a single-token chunk ($c_i = [t_j]$), its chunk representation $h_i^{c'}$ is the same as the token representation $h_j^{token}$. For a chunk with multiple tokens $(c_i = [t_{j}, \dots, t_{k}])$, the chunk representation $h_i^{c'}$ is the averaged token representations $(avg([h_{j}^{token}, \dots, h_{k}^{token}])$.
Moreover, we encode chunk type embedding with a trainable chunk type embedding $\boldsymbol{W_{type}}$ for additional type information:
\begin{equation}\label{eq:h_i_chunk}
    h_i^{c} = h_i^{c'} + \boldsymbol{W_{type}}(chunktype(c_i))
\end{equation}
where the function $chunktype(\cdot)$ returns the type (e.g. Noun, Verb) of the input chunk.

\paragraph{Dependency Graph Encoder.}
Given the sentence represented in SaC, we use GCN to encode the dependency structure of input sentence at chunk level.
For this purpose, we convert a token-level dependency structure to that of a chunk level by ignoring intra-chunk dependencies and retaining inter-chunk dependencies (see Figure~\ref{fig:dep_example}).

The chunk-level dependency graph is formulated as $G=(C,E)$, where the nodes in $C$ correspond to chunks $[c_1, \dots, c_m]$ and $e_{ij}$ in $E$ equals to 1 if there is a dependency between a token in node $c_i$ and a token in node $c_j$; otherwise, 0. Each node $c_i \in C$ has a node type.
We label a node with the type of the dependency from the node to its parent node (implemented by the function \say{$deptype$}).
We compute the node type embedding $l_{i}=\boldsymbol{W_{dep}}(deptype(c_i))$
with a trainable matrix $\boldsymbol{W_{dep}} \in \mathbb{R} ^ {d_l \times N_{dep}}$, where $d_l$ is the embedding dimension and $N_{dep}$ is the total number of unique dependency relations. 
Subsequently, we use GCN to encode $G$ with representations as follows:
\begin{equation}\label{eq:h_dep}
    h_i^{dep} = \mathrm{ReLU}\Big(\sum_{j=1}^{m} \alpha_{ij}(h_j^{c} + \boldsymbol{W_l} \cdot l_j + \boldsymbol{b}) \Big)
\end{equation}
where $m$ refers to the total number of chunk nodes in $G$, $W_l  \in \mathbb{R} ^ {d_h \times d_l}$ is a trainable weight matrix for dependency type embeddings, and $b  \in \mathbb{R} ^ {d_h}$ is the bias vector. The neighbour connecting strength distribution $\alpha_{ij}$ is calculated as below:
\begin{equation}
    \alpha_{ij} = \frac{e_{ij} \cdot \mathrm{exp}\big( (m_i)^T \cdot m_j \big)}{\sum_{k=1}^{m}e_{ik} \cdot \mathrm{exp} \big( (m_i)^T \cdot m_k \big)}
\end{equation}
where $m_i = h_i^{p} \oplus l_{i}$, and $\oplus$ is concatenation operator.
In this way, node type and edge information are modelled in a unified way.

\paragraph{SaC-based OIE Learning.}
We aggregate representations from the chunk representations in Equation~\ref{eq:h_i_chunk} and the graph representations in Equation~\ref{eq:h_dep} for chunk-level sequence tagging.
Note that the gold labels are provided at token level, whereas our predicted labels are at chunk level. 
To enable evaluation of the generated chunk-level tuples against the token-level gold labels, we assign the predicted probability of a multi-token chunk to all its member tokens.
Finally, we minimize the cross-entropy loss, computed between the predicted and the gold OIE tags:
\begin{equation}\label{eq:loss_oie}
    L_{OIE} = - \sum^n_{i=1} \sum^c_{j=1} y_{i,j}^{oie} \mathrm{log}(p_{i,j}^{oie})
\end{equation}
where $y^{oie}$ is the gold label, $p^{oie}$ is the Softmax probability. $c$ is the number of OIE span classes.

\subsection{\mname Model Training}
\label{sec:model_training}
The training procedure of \mname is as follows.
The SaC (chunking) part is first trained with the loss specified in Equation~\ref{eq:loss_sac} and chunking datasets described in Section~\ref{ssec:res_sac}.
The model weights of SaC part are fixed once trained, and are frozen during the training of SaC-based OIE tuple extraction. The OIE extractor is trained with the loss specified in Equation~\ref{eq:loss_oie} and OIE datasets described in Section~\ref{sec:setups}. Note that it is impossible to train the SaC and OIE extractor together. There is no dataset containing both chunking labels and OIE labels of ground-truth, while the multitask training requires both labels in same dataset. That is why \mname consists of 2 stages of training, rather than end-to-end training.

\section{Experiments}

\begin{table}
\centering
\small
\begin{tabular}{ l|c}
 \toprule
 Chunking model on CoNLL2000 &  Chunk type $F_1$\\
 \midrule
  AT~\cite{yasunaga-etal-2018-robust} & 95.3 \\
  Flair~\cite{akbik-etal-2018-contextual}  & 96.7 \\
  MAT~\cite{chen-etal-2020-enhance} & 97.0 \\
  ACE~\cite{wang-etal-2021-automated} & 97.3 \\
  \midrule
 Ours (BERT+Multi-task) & 97.0 \\
 \bottomrule
\end{tabular}
\caption{Chunk type $F_1$ on CoNLL 2000 chunking dataset. 
 Detailed results in Appendix~\ref{ssec:sac_details}. }
\label{tab:conll_f1}
\end{table}

\begin{table}
\small
\centering
    \begin{tabular}{@{}l|cc@{}}
    \toprule
Chunking model on OIA dataset &Boundary $F_1$ &
Type $F_1$ \\     
    \midrule
    Rule-based \cite{sun-etal-2020-predicate}  &  82.4 &  -  \\
    Neural model \cite{wang-etal-2022-oie} &  88.5 &  85.3$^\dag$      \\
    \midrule
    Ours (BERT+Multi-task) & 90.9  & 87.1       \\
    \bottomrule
    \end{tabular}
\caption{Performance of chunking on OIA dataset. Note that \citet{wang-etal-2022-oie} report chunk boundary result only and state that 96.4\% of them are labelled with correct types. We hence estimate their chunk type $F_1$ (marked with $^\dag$) based on the given percentage.}
\label{tab:chunk_overall}
\end{table}

\subsection{Performance of SaC}\label{ssec:res_sac}

\paragraph{Sentence Chunking Datasets.} 
CoNLL-2000 Shared Task dataset by \citet{tjong-kim-sang-buchholz-2000-introduction} annotates 8,936 / 2,012 sentences for Train/Test sets, respectively.
Open Information Annotation (OIA)  v1.1 dataset by \citet{wang-etal-2022-oie} contains 12,543 / 2,002 / 2,077 examples for Train / Development / Test sets. 
Each OIA annotation is a sentence-graph pair.
We only utilize the graph nodes (\ie simple phrases) for the chunking task.

\paragraph{SaC Evaluation Metric.} We report  Precision / Recall / $F_1$ for both chunk boundary detection and chunk type classification. For chunk boundary detection, we consider exact boundary match between a predicted chunk and a gold chunk as correct. For chunk type classification, the chunk is counted correct if both the boundary and type are exactly matched.    
That is, chunk type is  meaningful only if its boundary is detected correctly.

\paragraph{Sentence Chunking Results}
Reported in Table~\ref{tab:conll_f1}, our SaC model is comparable to the state-of-the-art ACE model~\cite{wang-etal-2021-automated} on the CoNLL-2000 dataset, even though our model is much simpler. Specifically, our model only utilizes BERT, while ACE leverages multiple word embeddings and PLMs including GloVe, fastText, ELMo, BERT, XLM-R, and XLNet.
On OIA dataset (see Table~\ref{tab:chunk_overall}), our SaC model outperforms all the previous methods.
The detailed results of chunk boundary detection and type classification, by chunk length and types, are summarized in Appendix~\ref{ssec:sac_details}.

\subsection{\mname Setups}\label{sec:setups}

\paragraph{OIE Dataset.}
We conduct experiments on four datasets: the two LSOIE datasets~\cite{solawetz-larson-2021-lsoie}, CaRB~\cite{bhardwaj2019carb}, and BenchIE~\cite{gashteovski-etal-2022-benchie}.\footnote{More details and statistics for OIE train/test sets are listed in Appendix~\ref{ssec:oie_training}.}

LSOIE is a large-scale OIE dataset converted from QA-SRL 2.0 in two domains, \ie Wikipedia and Science. It is 20 times larger than the next largest human-annotated OIE data, and thus is reliable for fair evaluation. LSOIE provides $n$-ary OIE tuples in the (\textit{ARG}\textsubscript{0}, \textit{Relation}, \textit{ARG}\textsubscript{1}, \dots, \textit{ARG}\textsubscript{n}) format.
We use both datasets, namely LSOIE-wiki and LSOIE-sci, for comprehensive evaluation. 

CaRB dataset is the largest crowdsourced OIE dataset.
CaRB provides 1,282 sentences with binary tuples. 
The gold tuples are in the (\textit{Subject}, \textit{Relation}, \textit{Object}) format.

BenchIE dataset supports a comprehensive evaluation of OIE systems for English, Chinese, and German. BenchIE provides binary OIE annotations and gold tuples are grouped according to fact synsets. In our experiment, we use the English corpus with 300 sentences and 1,350 fact synsets.

\paragraph{Evaluation Metric.}
For LSOIE-wiki and LSOIE-sci datasets, we follow \citet{dong2022smile_oie} to use exact tuple matching. A predicted tuple is counted as correct if its relation and all its arguments are identical to those of a gold tuple; otherwise, incorrect.
For the CaRB dataset, we use the scoring function provided by authors~\cite{bhardwaj2019carb}, which evaluates binary tuples with token level matching, \ie partial tuple matching. The score of a predicted tuple ranges from 0 to 1.
For the BenchIE dataset, we also adopt the scoring function proposed by authors~\cite{gashteovski-etal-2022-benchie}, which evaluates binary tuples with fact-based matching. A predicted tuple is counted as correct if it exactly matches to one fact tuple; otherwise, incorrect.

\begin{table*}
\small
\centering
\begin{tabular}{l|ll|ll|ll|lll}
 \toprule
 \multirow{2}{*}{Models} &
 \multicolumn{2}{c|}{LSOIE-wiki} &
 \multicolumn{2}{c|}{LSOIE-sci} &
 \multicolumn{2}{c|}{CaRB} &
 \multicolumn{3}{c}{BenchIE}
 \\
  & $F_1$ & AUC & $F_1$ & AUC & $F_1$ & AUC & $F_1$ & $Pr$ & $Re$\\
 \midrule
 \textbf{Token-level OIE Systems} & & & & & &\\
  CopyAttention~\cite{cui2018neural} & 39.5$^\dag$ & 35.9$^\dag$ & 48.8$^\dag$ & 46.8$^\dag$ & 51.6$^\ddag$ & 32.8$^\ddag$ & 21.5 & 26.4 & 17.5\\
  IMoJIE~\cite{kolluru2020imojie}  & 49.2$^\dag$ & 47.5$^\dag$ & 58.7$^\dag$ & 55.8$^\dag$ & 53.5$^\ddag$ & 33.3$^\ddag$ & 18.4 & 38.3 & 12.1 \\
  CIGL-OIE~\cite{kolluru2020openie6} & 44.7$^\dag$ & 41.9$^\dag$ & 56.6$^\dag$ & 52.3$^\dag$ & \textbf{54.0}$^\ddag$ & \textbf{35.7}$^\ddag$  & 25.4$^\S$ & 31.1$^\S$ & \textbf{21.4}$^\S$\\
  BERT~\cite{solawetz-larson-2021-lsoie}  & 47.5$^\dag$ & 44.7$^\dag$ & 57.0$^\dag$ & 53.2$^\dag$ & 51.4$^\dag$ & 30.6$^\dag$ & 23.1 & 32.5 & 17.9\\
  BERT+Dep-GCN~\cite{dong2022smile_oie} & 48.7$^\dag$ & 47.9$^\dag$    & 58.1$^\dag$ & 55.3$^\dag$    & 52.5$^\dag$ & 32.9$^\dag$ & 25.1 & 35.3 & 19.5\\
  SMiLe-OIE~\cite{dong2022smile_oie} & 51.7$^\dag$ & \textbf{50.8}$^\dag$ & 60.5$^\dag$ & 57.2$^\dag$ & \underline{53.8}$^\dag$ & 34.9$^\dag$ & 25.7 & 37.5 & 19.6 \\
 \midrule
 \textbf{Chunk-level OIE Systems } & & & & & &\\
 SpanOIE~\cite{zhan2020span}  &  47.5 &- & 57.5   &- & 49.4$^\ddag$  &- & 23.4 & 38.1 & 16.9\\
 OIE@OIA~\cite{wang-etal-2022-oie} & - & - & - & - & 52.3$^*$ & 32.6$^*$ & - & - & - \\
 \mname (SaC-NP\textsubscript{short})  & 50.7  & 48.9       & 60.3  & 58.4  & 53.0 & 33.8 & 25.3   &  40.2 & 18.5 \\
  \mname (SaC-NP\textsubscript{long})  & 48.5  &  46.4   & 57.2  & 56.7  & 50.9 & 31.7 & 23.4   & 35.1  & 17.6 \\
 \mname (SaC-OIA-SP) & \underline{52.1} & \underline{50.4}    & \textbf{61.2} & \underline{60.1}  & 53.6 &  \underline{35.5} & \underline{26.7}   &  \underline{41.5} & 19.7 \\
 \mname (SaC-CoNLL) & \textbf{52.6} & 50.2 & \underline{60.8} & \textbf{60.2} & 53.2 & 34.7 & \textbf{26.9} & \textbf{42.0} & \underline{19.8} \\
 \bottomrule
\end{tabular}
\caption{Results on four OIE datasets (best scores in boldface and second best underlined).  Scores with special mark are from  \cite{kolluru2020openie6}$^\ddag$,  \cite{gashteovski-etal-2022-benchie}$^\S$,  \cite{wang-etal-2022-oie}$^*$,  \cite{dong2022smile_oie}$^\dag$.}
\vspace{-0.5em}
\label{tab:baseline_systems}
\end{table*}

\subsection{OIE systems for Comparison}\label{sec:baselines}

\paragraph{Token-level OIE systems.}
CopyAttention~\cite{cui2018neural} is the first neural OIE model which casts tuple generation as a sequence generation task.
IMOJIE~\cite{kolluru2020imojie} extends CopyAttention and is able to produce a variable number of extractions per sentence. It iteratively generates the next tuple, conditioned on all previously generated tuples.
CIGL-OIE~\cite{kolluru2020openie6} models OIE as a 2-D grid sequence tagging task and iteratively tags the input sentence until the number of extractions reaches a pre-defined maximum.
Another baseline, BERT~\cite{solawetz-larson-2021-lsoie}, utilizes BERT and a linear projection layer to extract tuples.
SMiLe-OIE~\cite{dong2022smile_oie} explicitly models dependency and constituency graphs using multi-view learning for tuple extractions.
BERT+Dep-GCN is a baseline used in \citet{dong2022smile_oie}, which encodes semantic and syntactic information using BERT and Dependency GCN encoder. It is the closest baseline to our \mname. The difference is that \mname encodes dependency at chunk level and the chunks partially reflect the sentence syntactic information.

\paragraph{Chunk-level OIE systems.}
SpanOIE~\cite{zhan2020span} enumerates all possible spans from a given sentence and filters out invalid spans based on syntactic rules. Each span is subsequently scored to be relation, argument, or not part of a tuple.
OIE@OIA \cite{wang-etal-2022-oie} is a rule-based system that utilizes OIA graph. As the nodes of OIA graph are simple phrases (\ie chunks), we consider OIE@OIA as a chunk-level OIE system. 
\mname is our proposed model that is based on SaC for tuple extraction.
To explore the effect of different chunks in SaC, we implement four variants:  \mname (SaC-NP\textsubscript{short}),  \mname (SaC-NP\textsubscript{long}), \mname (SaC-OIA-SP), and \mname (SaC-CoNLL). Implementation details are listed in Appendix~\ref{sec:impl_details}.

\subsection{Main Results} \label{ssec:main_results}
Experimental results in Table~\ref{tab:baseline_systems} show that \mname, in particular its Sac-OIA-SP and SaC-CoNLL variants, achieve state-of-the-art results on three OIE datasets: LSOIE-wiki, LSOIE-sci, and BenchIE. Meanwhile, their results on CaRB are comparable with baselines.
We evaluate the statistical significance of \mname against its token-level baseline based on their $F_1$'s (each experiment is repeated three times with different random seeds). The $p$-values for \mname (SaC-OIA-SP) and \mname (SaC-CoNLL) are 0.0021 and 0.0027, indicating both results are significant at $p$ < 0.01.

\textbf{Comparing to token-level system}: \mname surpasses its token-level counterpart BERT+Dep-GCN on all the four datasets. Note that both \mname and BERT+Dep-GCN rely on BERT and Dependency GCN encoder; the only difference is the input unit, \ie chunks for \mname and tokens for BERT+Dep-GCN. Consequently, we suggest using chunks is more suitable to OIE. We observe SMiLe-OIE is a strong baseline. It explicitly models additional constituency information and the multi-view learning is computational complex. Comparing to it, \mname is simple yet effective.  CIGL-OIE performs the best on CaRB dataset. It adopts coordination boundary analysis to split tuples with coordination structure, which well aligns with the annotation guidelines of CaRB dataset, but not with the guidelines of the LSOIE and BenchIE datasets. In \mname, SaC treats chunks with coordination (\eg \say{Lily and Jimmy}) as a single unit, resulting in poor scores in such cases. Except on CaRB, CIGL-OIE cannot generalize well to other datasets.

\textbf{Comparing to Chunk-level system}: \mname achieves better results than SpanOIE, indicating that SaC is more reasonable than enumerated spans. 
\mname (SaC-OIA-SP) and \mname (SaC-CoNLL) are the best performing chunk-level OIE systems.
They outperform the other variants of \mname (SaC-NP\textsubscript{short}, SaC-NP\textsubscript{long}), which fail to model other type of chunks than noun phrases. Note that OIE@OIA generates tuples with rules manually crafted for OIE2006 and CaRB datasets. Also, the authors have not released source code of their rules. Therefore, OIE@OIA cannot be evaluated on LSOIE-wiki, LSOIE-sci, and BenchIE.

\begin{table}
\centering
\small
    \begin{tabular}{l|cc|cc}
    \toprule
    \multirow{2}{*}{\mname} &
    \multicolumn{2}{c|}{LSOIE-wiki} &
    \multicolumn{2}{c}{LSOIE-sci}\\
      & $F_1$ & AUC & $F_1$ & AUC\\
    \midrule
    OIA-SP & 52.1 & 50.4        & 61.2 & 60.1 \\
    -- w/o Dep-GCN  & 51.3  & 50.2        & 59.0 & 57.8  \\
    -- w/o Chunk type & 50.7 &  49.8        & 59.7  &   58.1      \\
        \midrule
    CoNLL & 52.6 & 50.2 & 60.8 & 60.2 \\
    -- w/o Dep-GCN  & 52.0  &  49.6        & 58.4 & 58.7  \\
    -- w/o Chunk type & 50.4 & 49.1         & 59.8 &  58.5       \\
    \bottomrule
    \end{tabular}
\caption{Ablation study of \mname.}
\vspace{-1.0em}
\label{tab:ablation}
\end{table}

\subsection{Ablation Study}

We ablate each part of \mname (SaC-OIA-SP) and \mname (SaC-CoNLL), and evaluate the ablated models on LSOIE-wiki and LSOIE-sci. The results are reported in Table~\ref{tab:ablation}.
We first remove the dependency graph encoder. In this setting, chunking representation obtained in Equation~\ref{eq:h_i_chunk} is directly used for tuple extraction. Results show that removing chunk level dependencies decreases the performance of \mname, indicating the importance of chunk-level dependency relations. To explore the importance of chunk type, we ablate the chunk type embedding as described in Equation~\ref{eq:h_i_chunk}. Observe that this also leads to performance degradation.

\section{Conclusion}
We propose Sentence as Chunk sequence (SaC) as an intermediate layer for OIE tuple extraction. Before utilizing chunks for OIE, it is crucial to understand to what extent chunks align with OIE gold tuple spans. We report detailed statistics of 4 different chunk choices for this purpose. We also compare SaC against OIA and show that SaC is more feasible, flexible and adaptable intermediate layer than OIA. We then design a strategy that simplifies the dependency structure of sentence but yet captures OIE-relevant dependency relations at chunk level. With the aid of SaC and chunk-level dependencies, \mname achieves state-of-the-art results on several OIE datasets.
As sentence chunking is well-studied, the simple yet effective \mname offers a new way to re-look at the OIE task. To develop more effective OIE models on top of SaC is our key focus in future work.

\clearpage
\section*{Limitations}
The limitations of \mname are analyzed from three perspectives: SaC chunking errors, syntactic parsing errors, and multiple extractions issue.
\textbf{(1)} Both CoNLL-chunked phrases and OIA simple phrases suffer around 10\% boundary violations as shown in Table~\ref{tab:span_analysis} (under Recall analysis). Since we use SaC as intermediate layer for OIE and perform tagging at chunk level, the chunk boundaries become a hard constraint of the extracted tuples. Among these violations, we examine 100 examples of OIA simple phrases and find that 55\% of these violations are caused by chunking errors due to some complicated sentence structures. The rest is mainly caused by tuple annotation errors, meaning that all OIE systems will suffer from these annotation errors.
\textbf{(2)} \mname relies on the chunk-level dependency relations as additional syntactic knowledge. Therefore, \mname will inevitably suffer from the noises introduced by the off-the-shelf dependency parsing tools. Also, we use POS tagger to extract all verbs in the sentence as tuple relation indicators. It is reported that the POS tagger fails to extract 8\% of verbs that are suppose to be relation indicators \cite{dong2022smile_oie}. Therefore, the discrepancy between POS verbs and tuple relations may affect the OIE quality.
\textbf{(3)} Moreover, there are 6\% of relation indicators corresponding
to multiple tuple extractions (one verb leading to more than one tuple), while our system extracts up to one tuple per relation indicator.

\bibliography{anthology}
\bibliographystyle{acl_natbib}

\appendix

\section{Appendix}
\label{Sec:appendix}

\subsection{Implementation Details and Resources} \label{sec:impl_details}
We build and run our system with Pytorch 1.9.0 and AllenNLP 0.9.0 framework.
The experiments are conducted with RTX 24GB 3090 GPU and Intel$^\circledR$  Xeon$^\circledR$ W-2245 3.90GHz CPU. 
Each epoch takes roughly 20 minutes for training on a single RTX 24GB 3090 GPU.
We run each experiment with three random seeds and report the averaged results.
We use NLP toolkit spaCy\footnote{\url{https://spacy.io/}} to extract the POS tags and dependency relations for sentences. 
In addition, we obtain  constituency parsing results through Stanford CoreNLP\footnote{\url{https://stanfordnlp.github.io/CoreNLP/}} and use the noun phrases to create NP-chunked phrases as part of our phrase selection exploration.
The hidden dimension $d_h$ for BERT representation $h_i^{bert}$, chunked phrase representation $h_i^{p}$, and Dep-GCN graph representation $h_i^{dep}$ are all set to 768.
The hidden dimension $d_l$ for Dep-Encoder type embedding $l_i^{dep}$ is 400.


\begin{table*}
\small
\centering
\begin{tabular}{l|l}
\toprule
Dataset & Resource URL\\
\midrule
CoNLL-2000 \cite{tjong-kim-sang-buchholz-2000-introduction} &\url{https://www.clips.uantwerpen.be/conll2000/chunking/}\\
OIA dataset v1.1 \cite{wang-etal-2022-oie} &  \url{https://github.com/baidu-research/oix}\\
 LSOIE dataset \cite{solawetz-larson-2021-lsoie} & 
\url{https://github.com/Jacobsolawetz/large-scale-oie}\\
 CaRB dataset and scoring code \cite{bhardwaj2019carb} & \url{https://github.com/dair-iitd/CaRB}\\
 BenchIE and scoring code \cite{gashteovski-etal-2022-benchie} &
\url{https://github.com/gkiril/benchie}\\ \midrule
Model & Source code URL \\
\midrule
BERT(base-uncased)~\cite{devlin2018bert} & \url{https://huggingface.co/bert-base-uncased} \\
CopyAttention~\cite{cui2018neural} & \url{https://github.com/dair-iitd/imojie}\\
IMoJIE~\cite{kolluru2020imojie} & \url{https://github.com/dair-iitd/imojie}\\
SpanOIE~\cite{zhan2020span} &\url{https://github.com/zhanjunlang/Span_OIE}\\
CIGL-OIE~\cite{kolluru2020openie6} &\url{https://github.com/dair-iitd/openie6}\\
SMiLe-OIE~\cite{dong2022smile_oie} & \url{https://github.com/daviddongkc/SMiLe_OIE}\\
\bottomrule
\end{tabular}
\caption{Online resources for datasets and models.}
\label{tbl:resourceUrl}
\end{table*}

The datasets, and their corresponding scoring scripts if applicable, used in this study are listed in Table~\ref{tbl:resourceUrl}. The table also list the source code URLs of the baseline models. 
\subsection{Four Chunk Choices and Their Statistics} \label{ssec:chunk_stats}

\paragraph{CoNLL chunks.} The CoNLL-2000 chunking task defines 11 chunk types based on the syntactic categories of Treebank~\cite{bies1995bracketing}. 
We train our own CoNLL-style chunking model, as described in Section~\ref{sec:sent_chunk}.


\paragraph{OIA simple phrases (OIA-SP).} 
We use the OIA simple phrases of 6 types defined by \citet{wang-etal-2022-oie}.
We also train our own OIA-style chunking model, as described in Section~\ref{sec:sent_chunk}.


\paragraph{NP chunks.} In this scheme, the tokens of a sentence are tagged with binary phrasal types: NP and O, where O refers to the tokens that are not part of any noun phrases.
We notice that there often exists nested NP. Accordingly, we create two types of NP chunks, \ie NP\textsubscript{short} and NP\textsubscript{long}. For example, the phrase \say{Texas  music player} is a nested NP.
NP\textsubscript{long} will treat it as a single NP, whereas NP\textsubscript{short} will split \say{Texas} and \say{music player} as two NPs.
We use Stanford constituency parser to get NP chunks.


\paragraph{SpanOIE spans.} SpanOIE~\cite{zhan2020span} enumerates all possible spans of a sentence, up to 10 words. 
To reduce the number of candidate spans, it keeps only the spans in which each word is either the syntactic parent or a child of another word in the same span.

The number, and the average length of  the chunks and gold spans are listed in Table~\ref{tab:length}.

\begin{table}[t]
\centering
\small
\begin{tabular}{ l|r|r}
 \toprule
 Spans/Chunks & Number of Spans & Average Length\\
 \midrule
 Gold Spans &  76,176 & 4.40 \\
 \midrule
 CoNLL &  339,099 & 1.62 \\
 OIA-SP & 307,505 & 1.77 \\
 NP\textsubscript{short} & 335,939 & 1.53 \\
 NP\textsubscript{long} & 225,796 & 2.28 \\
 SpanOIE & 1,995,281 & 4.34 \\
 \bottomrule
\end{tabular}
\caption{Number and average length of gold tuple spans, proposed phrases for SaC, and SpanOIE spans.}
\label{tab:length}
\end{table}



\subsection{Chunk Boundary and Type Analysis}
\label{ssec:sac_details}

\begin{table*}[t]
\centering
\small
\begin{tabular}{ l|rccc||rccc}
 \toprule
 Dataset& \multicolumn{4}{c||}{CoNLL-2000 }&\multicolumn{4}{c}{OIA } \\
 $L$\textsubscript{Chunk} & \#Chunk & $Pr$ & $Re$ & $F_1$& \#Chunk & $Pr$ & $Re$ & $F_1$\\
 \midrule
 1 token &  19,414 & 98.5 & 98.1 & 98.3 & 11,201 & 92.8 & 92.8 & 92.8\\
 2 tokens &  6,267 & 97.3 & 97.4 & 97.3& 2,924 &  88.0 & 89.5 & 88.7\\
 3 tokens & 2,865 & 96.8 & 97.3 & 97.1 & 1,245 &  84.5 & 85.8 & 85.1\\
 4 tokens & 945 & 94.1 & 96.4 & 95.2 & 440 &  78.3 & 78.0 & 78.1\\
 5+ tokens & 541 & 98.7 & 90.0  & 94.1 & 421 &  95.7 & 69.4  & 80.4\\
  \midrule
 Overall & 30,032 & 97.8 & 97.7  & 97.8 & 16,231 & 90.4 & 91.4  & 90.9\\
 \bottomrule
\end{tabular}
\caption{Chunk boundary extraction accuracy by chunk length.}
\label{tab:chunk_bound}
\end{table*}

\begin{table*}
\small
\begin{subtable}[t]{0.48\textwidth}
\centering
\begin{tabular}{ l|rccc}
 \toprule
 Type\textsubscript{chunk} & \#Chunk & $Pr$ & $Re$ & $F_1$\\
 \midrule
 NP & 12,422 & 97.5 & 97.3 & 97.4\\
 VP & 4,658 & 96.7 & 96.8 & 96.7 \\
 PP & 4,811 & 98.4 & 98.9 & 98.7 \\
 ADVP & 866 & 88.0 & 96.0 & 87.0 \\
 SBAR & 535 & 94.1 & 95.9 & 95.0 \\
 ADJP & 438 & 84.8 & 93.1 & 94.0 \\
 PRT & 106 & 77.9 & 89.6 & 83.3 \\
 O & 6,180 & 97.7 & 97.0 & 97.4\\
  \midrule
 Total     & 30,032 & 97.1 & 97.0 & 97.0 \\
 \bottomrule
\end{tabular}
\caption{CoNLL-2000}
\label{tab:Conll_type}
\end{subtable}
\hfill
\begin{subtable}[t]{0.48\textwidth}
\centering
\begin{tabular}{ l|rccc}
 \toprule
 Type\textsubscript{chunk} & \#Chunk & $Pr$ & $Re$ & $F_1$\\
 \midrule
 Noun & 7,159 & 86.8 & 85.8 & 86.3\\
 Verbal & 3,673 & 83.2 & 86.3 & 84.7\\
 Prepositional & 1,517 & 91.7 & 92.5 & 92.1\\
 Logical & 811 & 75.2 & 86.9 & 80.7\\
 Modifier   & 336 & 75.9 & 75.0 & 75.5\\
 Function  & 60 & 37.8 & 70.0 & 49.1\\
 O & 2,675 & 96.5 & 88.3 & 92.3 \\
  \midrule
 Total  & 16,231  & 86.6 & 87.5 & 87.1 \\
 \bottomrule
\end{tabular}
\caption{OIA dataset}
\label{tab:OIA_type}
\end{subtable}
\caption{Accuracy of chunk type classification by chunk type. Note that, for CoNLL-2000 datasets, CONJP, INTJ, LST and UCP each has fewer than 10 chunks, hence are excluded from the results.}
\end{table*}

Table~\ref{tab:chunk_bound} reports the chunk boundary accuracy of our SaC model (Section~\ref{sec:sent_chunk}) by chunk length in number of tokens. Observe that the $F_1$ of chunk boundary decreases when chunks become longer on both datasets. As expected, the longer the chunks, the harder the boundary detection becomes. Nevertheless, the $F_1$ of long chunk (\eg more than 5 tokens) is 94\% and 80.44\% on CoNLL-2000 and OIA datasets, respectively. This shows that our chunking model performs reasonably well in detecting long chunks. On the other hand, short chunks (\eg with 1 or 2 tokens) dominate the numbers, leading to high overall accuracy. We observe that the annotated sentences in CoNLL-2000 are longer and more formally written than that in OIA dataset. This could be a reason contributing to the higher $F_1$ on the CoNLL-2000 dataset.  

Recall that chunk type classification is conditioned on the boundary provided, \ie type is meaningful only if boundary is correctly detected. If the ground truth chunk boundaries are known, the overall type classification $F_1$ is 99.2\% and 95.8\% respectively, on  CoNLL-2000 and OIA datasets. However, in reality, the chunk boundaries have to be detected as well.

Tables~\ref{tab:Conll_type} and~\ref{tab:OIA_type} report the $F_1$ of chunk type classification by the major chunk types in both datasets. In this set of experiments, the chunk boundaries are detected together as type classification (\ie the same setting as in Section~\ref{sec:sent_chunk}). In both datasets, noun, verbal, and prepositional phrases dominate the chunks. The $F_1$ scores are reasonably high on these major types. Again, as the sentences in CoNLL-2000 datasets are much longer, the number of chunks in CoNLL-2000 is much larger than that in OIA dataset, although the two datasets have comparable number of test sentences.

\subsection{Chunk-level Dependency Modelling} \label{ssec:dep_chunk}

We argue that SaC simplifies the modeling of sentence syntactical structure. We elaborate this point with the example sentence shown in Figure~\ref{fig:dep_word}. In this sentence, \say{\textit{Lee}} is the appositional modifier (`appos') of \say{\textit{headmaster}}. However, it is actually the phrase \say{\textit{Ms. Lee}} that is appositional to the phrase \say{\textit{the headmaster}}. If we want to model the relation between \say{\textit{Ms.}} and \say{\textit{the}} through token dependencies, we need to pass through three hops (`compound' $\rightarrow$ `appos' $\rightarrow$ `det') in order to link them up. In contrast, connecting \say{\textit{Ms.}} and \say{\textit{the}} via chunk-level dependencies only requires a single hop (`appos'). In another case, \say{\textit{Lee}} is the nominal subject (`nsubj') and \say{\textit{Lily}} is the direct object (`dobj') of verb \say{\textit{told}}. Apparently, we need additional dependency relations to locate the complete subject and object of \say{\textit{told}}. 
If we model dependencies at chunk level, the complete subject and object of \say{\textit{told}} can be easily located to be  \say{\textit{Ms. Lee}} and \say{\textit{Lily and Jimmy}} respectively. 

The conversion to chunk-level dependency relations from token-level is performed in two steps. \textbf{(1)} We remove a dependency relation between two tokens if both tokens are within the same chunk. The following dependency relations in Figure~\ref{fig:dep_word} are removed: \say{compound} relation between \say{\textit{Ms.}} and \say{\textit{Lee}}, \say{det} relation between \say{\textit{the}} and \say{\textit{headmaster}}, \say{cc} relation between \say{\textit{Lily}} and \say{\textit{and}}, \say{conj} relation between \say{\textit{Lily}} and \say{\textit{Jimmy}}, and \say{acomp} relation between \say{\textit{is}} and \say{\textit{responsible}}. \textbf{(2)} We map the remaining dependency relations, that are between tokens from different chunks, to be the relations between chunks. For example, the \say{appos} relation between \say{\textit{Lee}} and \say{\textit{headmaster}} is map to \say{\textit{Ms. Lee}} and \say{\textit{the headmaster}} as shown in Figure~\ref{fig:dep_phrase}. 
Similarly,  \say{\textit{Ms. Lee}} turns into the nominal subject (nsubj) and \say{\textit{Lily and Jimmy}} becomes the direct object (dobj) of verb \say{\textit{told}}.

\subsection{Details of OIE Datasets} \label{ssec:oie_training}
In this section, we elaborate more details about the train/test set of OIE datasets as mentioned in Section~\ref{sec:setups}.
For LSOIE, we follow \citet{solawetz-larson-2021-lsoie} and \citet{dong2022smile_oie} to split the train/test set in LSOIE-wiki and LSOIE-sci domain, respectively. The statistics of LSOIE train/test sets are listed in Table~\ref{tab:data_set}.

CaRB only provides 1,282 annotated sentences and BenchIE provides 300 sentences, which are insufficient for training neural OpenIE models.
As a result, we use the CaRB and BenchIE dataset purely for testing. We follow \citet{kolluru2020openie6} to convert bootstrapped OpenIE4 tuples as labels for distant supervised model training. The statistics of CaRB and BenchIE train/test sets are listed in Table~\ref{tab:data_set}.

\begin{table}[t]
\small
\centering
\begin{tabular}{ l|crr}
 \toprule
 Dataset & Source & \#Sent & \#Tuple\\
 \midrule
 LSOIE-wiki-train & QA-SRL 2.0 & 19,591 & 45,890 \\
 LSOIE-wiki-test & QA-SRL 2.0 & 4,660 & 10,604 \\
 \midrule
 LSOIE-sci-train & QA-SRL 2.0 & 38,826 & 80,271 \\
 LSOIE-sci-test & QA-SRL 2.0 & 9,093 & 17,031 \\
  \midrule
 CaRB-train & OpenIE 4 & 92,774 & 190,661 \\
 CaRB-test & Crowdsourcing & 1,282 & 5,263 \\
  \midrule
 BenchIE-train & OpenIE 4 & 92,774 & 190,661 \\
 BenchIE-test & Expert & 300 & 1,350 \\
 \bottomrule
\end{tabular}
\vspace{-0.5em}
\caption{Statistics of OIE datasets used in training and evaluating \mname.}
\vspace{0.5em}
\label{tab:data_set}
\end{table}

\end{document}